\documentclass[sn-basic]{sn-jnl}% Basic Springer Nature Reference Style/Chemistry Reference Style
\usepackage{graphicx}%
\usepackage{multirow}%
\usepackage{amsmath,amssymb,amsfonts}%
\usepackage{amsthm}%
\usepackage{mathrsfs}%
\usepackage[title]{appendix}%
\usepackage{xcolor}%
\usepackage{textcomp}%
\usepackage{manyfoot}%
\usepackage{booktabs}%
\usepackage{algorithm}%
\usepackage{algorithmicx}%
\usepackage{algpseudocode}%
\usepackage{listings}%
\begin{document}

\title[Artificial Intelligence and Human Geography
]{Artificial Intelligence and Human Geography \\ \\
\small Chapter in the Encyclopedia of Human Geography }

\author{\fnm{Song} \sur{Gao}}\email{song.gao@wisc.edu}

\affil{\orgdiv{Geospatial Data Science Lab, Department of Geography}, \orgname{University of Wisconsin-Madison}, \orgaddress{\street{550 N Park Stret}, \city{Madison}, \postcode{53706}, \state{WI}, \country{USA}}}

\abstract{ 
This paper examines the recent advances and applications of AI in human geography especially the use of machine (deep) learning, including place representation and modeling, spatial analysis and predictive mapping, and urban planning and design. AI technologies have enabled deeper insights into complex human-environment interactions, contributing to more effective scientific exploration, understanding of social dynamics, and spatial decision-making. Furthermore, human geography offers crucial contributions to AI, particularly in context-aware model development, human-centered design, biases and ethical considerations, and data privacy. The synergy beween AI and human geography is essential for addressing global challenges like disaster resilience, poverty, and equitable resource access. This interdisciplinary collaboration between AI and geography will help advance the development of GeoAI and promise a better and sustainable world for all.
}



\maketitle

%\textbf{Synonyms: AI, Robotics}

\section{Introduction}\label{sec:intro}
Artificial intelligence (AI) is a field in computer science and engineering that focuses on developing intelligent machines capable of performing problem-solving tasks and achieving goals that typically require human intelligence~\citep{turing1950computing,mccarthy2004artificial}. 
There are generally two types of AI: \textit{Weak AI} and \textit{Strong AI}.  \textit{Weak AI} sits at the foundation of AI development, with its focused human-like specific abilities (e.g., conversation in Siri) and practical applications (e.g., facial recognition from images and videos) in specific contexts, while \textit{Strong AI} (known as Artificial General Intelligence) represents the ultimate goal of AI research, aiming to replicate or even surpass human intelligence to solve problems, learn, and plan for the future with self-aware consciousness~\citep{goertzel2014artificial}.

AI is revolutionizing many domains including geographical sciences, presenting both immense opportunities and formidable challenges. The rapid advancement of AI is fueled by theoretical breakthroughs in neural science and computer science, the ubiquity of big data, advanced computer hardware such as graphics processing units (GPUs) and powerful high-performance computing platforms that enable the efficient development training, and deployment of AI models, and the flourishing AI products, services and applications that are changing human behaviors~\citep{gao2020review,li2020geoai,torrens2018artificial}.

As \cite{couclelis1986artificial} expressed her excitement about the AI-driven evolution---``It is not often that geography is touched by a development having the potential to affect substantially all of the practical, technical, methodological, theoretical and philosophical aspects of our work.'' The intersection of AI and geography has a rich and evolving history, with its early roots documented in works by~\cite{couclelis1986artificial,smith1984artificial,openshaw1997artificial}.  Prior to the surge of deep learning research in the 2010s~\citep{lecun2015deep} and its applications in geography and earth sciences~\citep{hu2019artificial,goodchild2021replication,liu2022review,reichstein2019deep}, significant AI advancements included theoretical explorations in the 1950s and 1960s~\citep{buchanan2005very}; the emergence of artificial neural networks, heuristic search algorithms, knowledge-based symbolic expert systems, neurocomputing, and artificial life concepts (e.g., cellular automata) in the 1980s; the development of genetic programming, fuzzy logics, and hybrid intelligent systems in the 1990s~\citep{openshaw1997artificial}; the integration of ontology and semantics for information retrieval and knowledge graphs in the 2000s; and the recent development of foundation model \citep{bommasani2021opportunities,mai2023opportunities}. All of these advances have laid the foundation for the burgeoning field of Geospatial artificial intelligence (GeoAI)~\citep{gao2023handbook}.

Rooted in geography and geographic information science (GIScience), GeoAI has now emerged as a transformative interdisciplinary field, integrating geographical studies with AI techniques, particularly spatially-explicit machine learning and deep learning methods as well as geo-knowledge graphs~\citep{gao2020review,janowicz2020geoai,mai2022symbolic}. This convergence has spurred groundbreaking advancements in both academia and industry. GeoAI encompasses the development of intelligent computer programs that emulate human perception, spatial reasoning, and the ability to uncover insights from geographical phenomena and understanding their complex dynamics through coupling spatial knowledge and earth science process-based models with deep neural networks~\citep{gao2021geospatial,reichstein2019deep}. The researches in GeoAI can deepen our understanding of complex human-environmental systems and their interactions, particularly within a spatial context. 

Regarding the topics of AI in human geography, a recent conversation about GeoAI, counter-AI, and human geography by \cite{janowicz2022geoai} discussed opportunities and challenges in defining intelligence for machines, and the role of humans in AI development and decision-making processes, emphasizing the need for ethical considerations and accountability in the development of GeoAI. In the following, I will first introduce some of the recent advances of AI in human geography and then discuss the role of human geography in AI development.  

\section{AI in Human Geography}\label{sec:AIHuman}

Human geography has a long tradition of studying the spatially and temporally varying interrelationships between people, place, and environment, and the ways in which people and their activities shape, and are shaped by, the physical, economic, political, and cultural environment~\citep{jones2012human}. The integration of AI technologies has significantly expanded the capacity of researchers and practitioners in understanding the complex human-environment interactions. Geographers have explored a wide range of topics and made advances in the following areas (but not limited to).

\textbf{Place Representation and Modeling: }
Place can serve as a function between location and people~\citep{mennis2016modeling}, a function of location, activity and time~\citep{mckenzie2017juxtaposing}, and a function of social relations~\citep{giordano2018limits}. Traditionally, data about places were collected through mapping agencies (e.g., digital gazetteers) and survey-based narratives. The emergence of geospatial big data brings new opportunities to extract fine spatiotemporal resolution of human-place interaction data and understand the rich place semantics from large-scale volunteered geographic information and crowdsourced data streams, such as social media posts (including texts, photos, and videos), GPS tracks, text reviews on points of interest (POIs) and neighborhoods, and other Web documents~\citep{gao2017data,gao2017constructing,liu2015social}. A wide-range of AI methods provide new opportunities to understand and extract the characteristics of places as well as associated human activities, experiences, emotions, and movements in different contexts. 
For example, urban areas of interest that attract people's attention but with different spatial patterns and region-specific semantics were automatically extracted from georeferenced photos posted on social media by employing spatial and spectral clustering algorithms, and natural language processing techniques~\citep{hu2015extracting}. In addition, the spatial and hierarchical semantics between places that support human cognition of places, identification of urban functional regions, and information retrieval in digital maps, were modelled using POI embeddings (i.e., multidimensional vectors) via topic modeling~\citep{gao2017extracting} and deep learning techniques~\citep{huang2022estimating}. Patterns and relations between places can be computed and extracted from collective human descriptions using place graphs~\citep{chen2018georeferencing}. Empowered by the state-of-the-art human face and emotion recognition AI techniques, human emotions at different places were extracted from millions of georeferenced photos~\citep{kang2019extracting}, which enabled the study of spatially embodied emotions in human geography~\citep{simonsen2007practice}. Recently, researchers found that geo-knowledge-guided large language models (e.g., ChatGPT) improved the extraction of location textual descriptions from disaster-related social media messages, which can facilitate collaborations between disaster response experts and AI developers and ultimately help save human lives~\citep{hu2023geo}.

\textbf{Spatial Analysis and Predictive Mapping: }
AI has revolutionized geospatial data analysis, interpretation, and modeling of human-environment relations by providing innovative spatial analytical tools and methodologies. New or modification of spatial analysis methods and spatial statistical models have been proposed by incorporating deep neural networks, which we termed as Intelligent Spatial Analytics~\citep{zhu2022towards}. Geographically weighted artificial neural network (GWANN)~\citep{hagenauer2022geographically} and  geographically neural network weighted regression (GWNNR)~\citep{du2020geographically} models were developed by constructing the spatial nonstationary weights with neural networks and feeding into  the geographically weighted regression for more accurate predictive modeling. 
Similarly, new spatial regression graph convolutional neural networks (SRGCNNs) were developed for the spatial regression analysis of geographical multivariate distributions and outperformed traditional spatial autoregressive models~\citep{zhu2021spatial}. 
Machine learning and deep learning algorithms have also been widely used for data-driven modeling and predictive mapping in geography~\citep{miller2015data} and identifying patterns and relationships that would be nearly impossible for human researchers to discern manually~\citep{lavallin2021machine}. For example, \cite{zhu2020understanding} applied the graph convolutional neural networks (GCNNs) for modeling place graphs and inferring the unknown properties of a place with a high prediction accuracy by utilizing both the observed attributes (e.g., visual and functional features) and the relational characteristics (e.g., distance, adjacency, spatial interaction intensity) of the places.  When applying state-of-the-art deep learning models for population mapping in geography, the issues on model selection, neighboring effects, and systematic biases play an important role on the predictive mapping accuracy~\citep{huang2021sensing}. Researchers found that population estimation performance reduced with increasing neighborhood sizes and a pervasive bias existed in which all tested deep learning models (i.e., VGG, ResNet, Xception, and DenseNet) overestimated sparsely populated patches and underestimated densely populated ones~\citep{huang2021sensing}. This work also demonstrated the importance of spatial concepts and some limitations when applying AI methods in human geography due to spatial heterogeneity. Future efforts need to be made towards the (weak) replication of GeoAI models across space and time in the social and environmental sciences~\citep{goodchild2021replication}.

\textbf{Urban Planning and Design: }  
Human geographers have long been interested in understanding the intricate relationships between people and the environment and collaborating with urban planners for planning our cities. By applying AI algorithms, they can effectively interpret large geospatial datasets, including census surveys, satellite imagery, street-view images, social media posts, and other sensor data to support urban analytics~\citep{de2023geoai}. By utilizing the state-of-the-art deep learning models, human perception of neighborhood playability for childhood development in cities~\citep{kruse2021places} and urban visual intelligence about the hidden neighborhood socioeconomic status, such as poverty status and health outcomes, can be extracted from street-view images~\citep{fan2023urban,kang2020review}; near real-time global land cover and land use patterns and temporal changes can be detected from 10m spatial resolution of Sentinel-2 remote sensing imagery~\citep{brown2022dynamic}. Urban geographers and planners can also extract valuable insights into population dynamics, traffic patterns, land use patterns, environmental changes, and the interactions between social and ecological systems, so as to optimize urban spatial structure, improving transportation efficiency and quality of living, and environmental sustainability~\citep{mortaheb2023smart,liu2022review}. AI-driven predictive modeling can assist in rooftop solar potential estimation for sustainable city design~\citep{wu2021roofpedia}, and anticipating changes in population density and mobility, facilitating better resource allocation and disaster preparedness in ~\citep{zou2022empowering,vongkusolkit2021situational}. 
Additionally, AI-powered semantic and sentiment analyses of online neighborhood reviews helped better understand the perceptions and emotions of people toward their living neighborhoods and environments, which can be applied for supporting urban planning and improving quality of life studies~\citep{hu2019semantic}.
Furthermore, the advances in generative AI models brings exiting opportunities for urban planners to facilitate the automatic rendering of urban master plans~\citep{ye2022masterplangan} and building floorplan layouts~\citep{wu2022generative} via generative adversarial networks (GANs), otherwise they need to rely on subjective design and labor-intensive production process. However, the generative AI also introduces the deep fakes into geography~\citep{zhao2021deep} and we need to make more efforts on the investigation of trustworthiness of such AI generated geospatial data and city plans.

\section{Human Geography for AI}\label{sec:HumanForAI}
Human geography also contributes to the development and applications of AI by providing valuable insights into the social, cultural, and economic factors that shape human behavior and interactions with the technology.

\textbf{Context-Aware and Debiasing AI: }  
Human geography contributes to AI through the development of context-aware models and the analysis of biases of AI models when applied in heterogeneous contexts and places. Human perception of the physical and socioeconomic environments through deep learning has attracted large interest of scholars from geography, computer sciences, urban planning, and environmental psychology~\citep{gebru2017using,fan2023urban,kang2020review,han2022measuring}. Human perception of places reflects people’s subjective feelings towards their surrounding environment, e.g., whether a place is safe, lively, and wealthy~\citep{zhang2018measuring}. However, there exist multiple types of biases in existing AI-driven computational frameworks when quantifying human perceptions of places: \textit{data bias, model bias, and perception bias}. \cite{kang2023assessing} compared the safety perception measures from the survey based on neighborhood residents’ responses with those from the GeoAI approach with street-view imagery to better understand the relationship between the two types of measures. This research found that citywide residents, but not neighborhood residents, may feel economically vibrant places look safe; elder people may underestimate the safety of their living places which may enlarge perception bias. Therefore, integrating cultural, historical, and socioeconomic insights into pre-trained AI models and fine-tuning them with local data and participant inputs may enhance AI's ability to comprehend and respond to spatial heterogeneity and diverse human experiences. This approach not only improves the accuracy of AI applications but also fosters inclusivity and respect for social diversity across geographic regions. 

\textbf{Data Privacy and Ethical Framework: }  
Human geography addresses ethical concerns related to data collection and usage, as well as community trust~\citep{rowe2021big}. This knowledge is invaluable for AI developers who must navigate the ethical implications of collecting and analyzing vast amounts of data regarding privacy and security. In GeoAI development, such concerns often refer to the use or exposure of sensitive geospatial information such as a user's home location, workspace, POI preferences, daily trajectories, and associated personal inferences based on such information~\citep{kessler2018geoprivacy,rao2023cats}. There is a trade-off between the data utility in downstream tasks (e.g., user profiling) and user privacy~\citep{gao2019exploring}. In human mobility studies, common user privacy protection methods include \textit{de-identification, geomasking, trajectory k-anonymization, and differential privacy}. Recently, deep neural networks such as LSTM and GANs have been used in human movement trajectory generation tasks with the aim to balance the data utility-privacy trade-off~\citep{rao2020lstm,rao2023cats}. With the substantial advances in AI foundation
models, recent studies, however, revealed that the development and use of foundation models could potentially unveil substantial privacy and security risks, including the disclosure of sensitive information, representational bias, hallucinations, and misuse~\citep{bommasani2021opportunities}. In the lifecycle of building and using GeoAI foundation models~\citep{mai2023opportunities}, a series of potential privacy and security risks that existed around the pre-training and fine-tuning stages with geospatial data, centralized serving and tooling, prompting-based interaction, and feedback mechanisms~\citep{rao2023building}.
Looking ahead, human geographers and GIScience scholars can further contribute to establishing guidelines that ensure responsible and privacy-conscious GeoAI practices.

\textbf{Human-Centered Design: }  
Understanding the perspectives, experiences, emotions, and lived realities of people in different places and regions is essential for human geography research~\citep{tuan1979space}. Therefore, human geography emphasizes the importance of human-centered design in place-based innovation and policy making, aligning seamlessly with the principles of ethical (Geo-)AI development~\citep{siau2020artificial,kang2024artificial} and user-centered design in cartography~\citep{roth2017user}. By incorporating human geographers in the (Geo-)AI design processes, developers can create AI systems that better serve diverse communities and mitigate potential negative impacts on our society, which is  well aligned with the core idea of keeping ``human-in-the-loop'' for the development of AI systems and algorithms~\citep{janowicz2022geoai,zanzotto2019human}. Several principles and practices like collaborative engagement with data~\citep{lloyd2011human}, accountability mechanisms in AI systems design~\citep{mittelstadt2019principles} and empathy~\citep{srinivasan2022role} can be considered for addressing the social-technical challenges in the AI paradigm.

\section{Conclusion}\label{sec:conclusion}

The relationship between AI and human geography is not a one-way street. As AI becomes increasingly sophisticated, its applications in human geography will become even more diverse and impactful. At the same time, the abovementioned critical insights provided by human geographers will be crucial for ensuring that less-biased (Geo-)AI is developed and deployed in a responsible and ethical manner. Mitigating multiple types of biases throughout the ``data-model-action'' loop in human-environment systems may ensure equitable outcomes and AI-driven spatial decision making.
By working together, AI and human geography can address some of the most pressing challenges facing human-environment nexus today, including climate change, poverty, food security, health, disaster resilience, and equitable access to resources and services. This interdisciplinary collaboration has the potential to create a better and sustainable world for all.

\backmatter

\bibliography{AI-references} 

\end{document}